\documentclass[10pt,twocolumn,letterpaper]{article}

\usepackage{cvpr}
\usepackage{times}
\usepackage{epsfig}
\usepackage{graphicx}
\usepackage{amsmath}
\usepackage{amssymb}
\usepackage[numbers,sort]{natbib}
\usepackage{subfig}
\usepackage[skip=0pt]{caption}
\usepackage{appendix}

\usepackage[pagebackref=true,breaklinks=true,letterpaper=true,colorlinks,bookmarks=false]{hyperref}

\cvprfinalcopy 

\def\cvprPaperID{3730} 

\ifcvprfinal\fi
\begin{document}

\title{AVA-ActiveSpeaker: An Audio-Visual Dataset for Active Speaker Detection}

\author{Joseph Roth, Sourish Chaudhuri, Ondrej Klejch, Radhika Marvin, Andrew Gallagher, Liat Kaver,\\
Sharadh Ramaswamy, Arkadiusz Stopczynski, Cordelia Schmid, Zhonghua Xi, Caroline Pantofaru\\
\\
Google AI Perception\\
{\tt\small \{josephroth, sourc, klejcho, radahika, agallagher, lkaver,}\\
{\tt\small sharadh, astopczynski, cordelias, zxi, cpantofaru\}@google.com}
}

\maketitle

\begin{abstract}
    Active speaker detection is an important component in video analysis algorithms for applications such as speaker diarization, video re-targeting for meetings, speech enhancement, and human-robot interaction. The absence of a large, carefully labeled audio-visual dataset for this task has constrained algorithm evaluations with respect to data diversity, environments, and accuracy. This has made comparisons and improvements difficult. In this paper, we present the AVA Active Speaker detection dataset (AVA-ActiveSpeaker) that will be released publicly to facilitate algorithm development and enable comparisons. The dataset contains temporally labeled face tracks in video, where each face instance is labeled as speaking or not, and whether the speech is audible. This dataset contains about 3.65 million human labeled frames or about 38.5 hours of face tracks, and the corresponding audio.  We also present a new audio-visual approach for active speaker detection, and analyze its performance, demonstrating both its strength and the contributions of the dataset.
\end{abstract}

\section{Introduction}
\label{sec:introduction}

\begin{figure}[t]
  \centering
  \includegraphics[width=\linewidth]{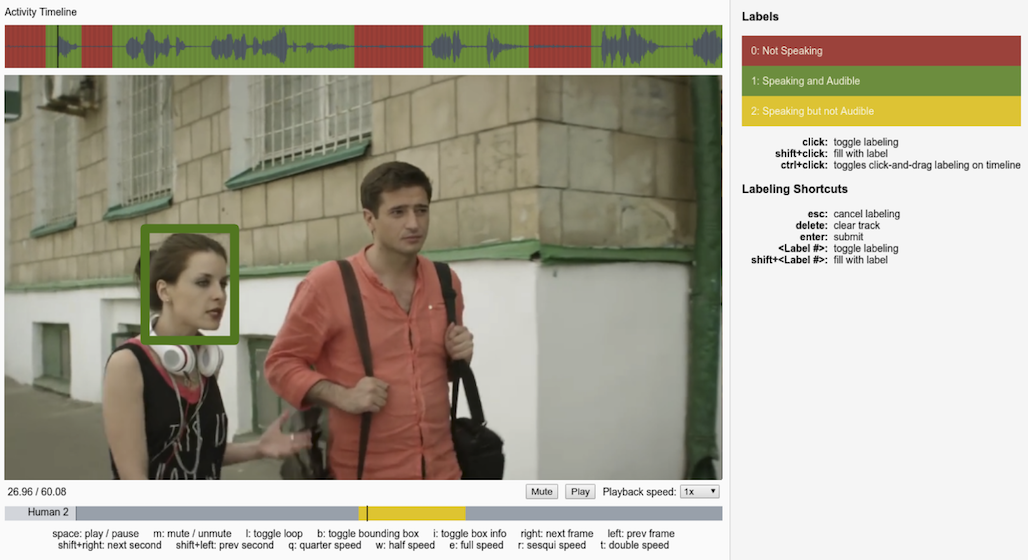}
  \caption{The annotation interface for AVA-ActiveSpeaker. Given its surrounding video and audio (waveform visualized above the frame), each face is annotated with whether it is speaking and whether the speech is audible. Annotations are continuous in time.  Details in Sec.~\ref{sec:labeling-speechannotation}. Active speaker detection involves classifying a given face at a given time as speaking or not.}
  \label{fig:rating_ui}
  \vspace{-2mm}
\end{figure}

Conversational content in videos has received significant attention in the literature, with audio-only, visual-only, and joint audiovisual modeling approaches applied to applications such as human-robot interactions, speech recognition and analysis, and video re-targeting. Active speaker detection--- detecting which (if any) of the visible people in a video are speaking at any given time--- is a core component in many of these applications. In this paper, we present the AVA Active Speaker detection dataset (AVA-ActiveSpeaker) as a benchmark dataset, which we will release publicly. Alongside the dataset, we present a state-of-the-art audiovisual algorithm for active speaker detection and a detailed analysis of its performance.

Active speaker detection has multiple applications, such as in interactive systems that identify the speaker and personalize responses \cite{stefanov2017vision, stefanov2016look}, in speech transcription, speaker diarization and speech enhancement systems \cite{chung2016out, shillingford2018lsvsr, shinoda2011speaker, hoover2018putting, afouras18conversation, ephrat2018looking}, and for tracking storylines and characters in narrative content \cite{cour2008movie, everingham2009taking}. It is also used to facilitate the mining of training data for modeling these tasks \cite{shillingford2018lsvsr, ephrat2018looking, nagrani2018voxceleb}. When active speaker detection is used to mine task-relevant data, prior work often takes an approach biased towards high precision when finding video clips of likely speakers. As a result, many of the trained models have not been exposed to difficult data, and are at risk of failing to generalize to real-world applications.

The challenges to robust active speaker detection modeling come from two sources. 
The first is the intrinsic difficulty of the task. Visual-only approaches are confused by other face/mouth motions: eating, expressions, holding a hand up to the mouth, or yawning. Audio-only speech detection cannot be associated with a visual person detection without constraining assumptions ({\it e.g.} speaker is always visible) that do not generalize.
The second is that web video content comes from diverse demographics, recording device resolution, contain occlusions and varied illumination settings. 
Robust modeling for active speaker detection requires joint audiovisual models trained from a large and diverse dataset.

Such a dataset did not previously exist. The closest related work is that of Chakravarty {\em et al.}~\cite{chakravarty2015who,chakravarty2016active, chakravarty2016cross}, and their entire experimental data includes seven $30$-min PhD thesis presentations and segments of a single video with a panel discussion from YouTube. We expect the broader community to benefit significantly from scaling up the training and evaluation corpus to a much larger and more diverse dataset.

The AVA-ActiveSpeaker dataset fills this need. Videos in the dataset are from the diverse AVA v1.0 action recognition dataset of YouTube movies~\cite{gu2018ava}. Given the audio and video, each face in each frame is labeled as speaking or not, and whether the speech is audible. Fig.~\ref{fig:rating_ui} shows the annotation interface, with dense temporal annotation of the highlighted face shown on the waveform above. The dataset contains about 3.65 million labeled frames, about 38 hours of face tracks, and the corresponding audio.

\begin{figure}[!t]
\captionsetup[subfigure]{labelformat=empty}
   \centering
   \subfloat[][]{\includegraphics[width=.45\linewidth]{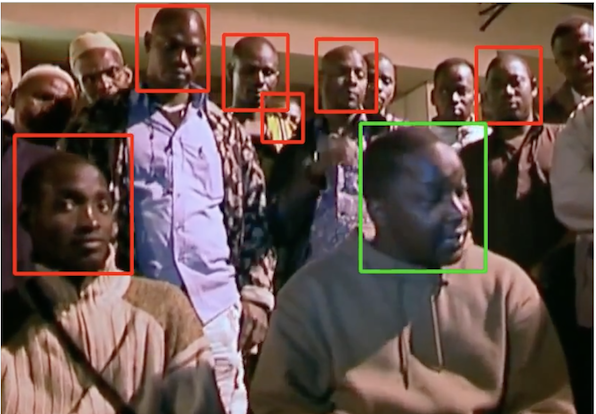}}\quad
   \subfloat[][]{\includegraphics[width=.45\linewidth]{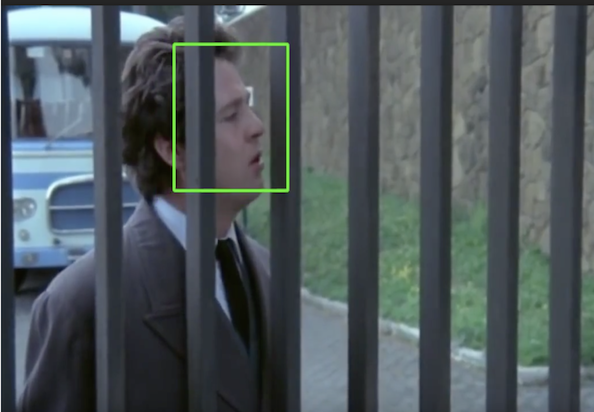}}\\
   \vspace{-1.8\baselineskip}
   \subfloat[][]{\includegraphics[width=.45\linewidth]{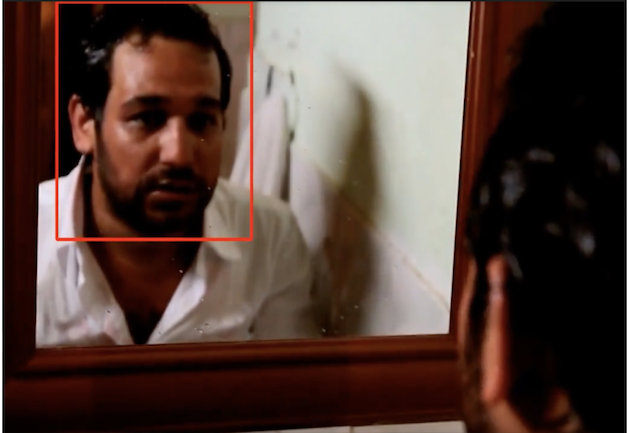}}\quad
   \subfloat[][]{\includegraphics[width=.45\linewidth]{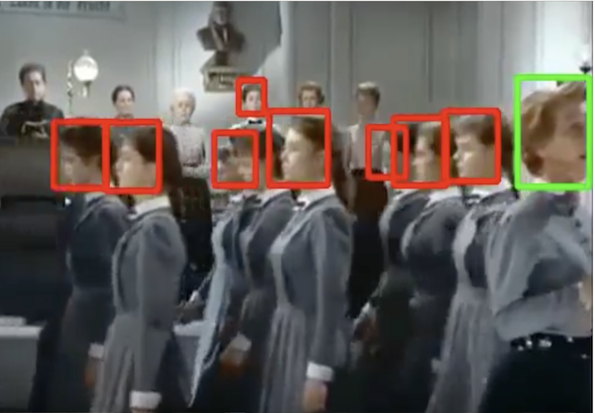}}
   \caption{Examples of labeled faces in AVA-ActiveSpeaker. A green box implies ``Speaking and Audible" label, while red implies ``Not Speaking".}
   \label{fig:sample_labels}
   \vspace{-2mm}
\end{figure}

We describe the process of obtaining annotations in detail in Section \ref{sec:labelinginterface}, and present an analysis of the annotated data, including its relationship to past labels released on AVA in Section \ref{sec:datasetstatistics}. A sample of labeled face frames is shown in Figure \ref{fig:sample_labels}, which includes partial occlusion, a variety of face sizes, activities, demographics and lighting conditions.

Alongside the dataset, we present an audiovisual model for active speaker detection in Section \ref{sec:approach}. The model is real-time, and is trained directly from the pixels and audio without any pre-trained embeddings. A detailed analysis of model performance is presented in Section \ref{sec:evaluation}. The analysis shows the value of the label ecosystem on the AVA corpus; speech activity labels from AVA-Speech~\cite{chaudhuri2018avaspeech} allow us to evaluate performance in the presence of different background noise conditions, the temporal extent of labels in AVA-ActiveSpeaker enable fully supervised evaluation of recurrent models, and the multimodal nature of the dataset enables broader model exploration. 

We summarize our contributions as follows:
\begin{itemize} 
\itemsep0em
\item A large-scale, human-annotated, diverse, end-application agnostic, public benchmark dataset, with dense, spatio-temporal labels for active speaker detection. This additionally builds upon the increasingly rich ecosystem of labels on the AVA corpus, enabling deeper analysis and label sharing across tasks.
\item A real-time, joint audiovisual model for this task, end-to-end trained directly off of the pixels and audio without the use of any pre-trained networks.
\item State-of-the-art benchmark results for various models, along with a careful analysis of model performance, and the effect of conditions that inform downstream application-specific modeling choices.
\end{itemize}
\section{Related work}
\label{sec:relatedwork}

In this section, we group related literature into 3 broad groups--- applications that make use of active speaker detection, prior work on developing datasets, and prior work with multimodal modeling.

\textbf{Applications using active speaker detection:}
The active speaker detection module is often handled via heuristics in the context of larger end-applications. For example: Everingham {\em et al.}~\cite{everingham2006buffy} assume motion in the lip area implies speech, Chung {\em et al.}~\cite{chung2016out} and Nagrani {\em et al.}~\cite{nagrani2018voxceleb} assume a single visible face is the speaker, and Shillingford {\em et al.}~\cite{shillingford2018lsvsr} use a combination of both. All of these works involve applications tolerant to biasing the active speaker detection module toward precision, and the last 3 use it to mine and label datasets. As discussed in Sec.~\ref{sec:introduction}, this may result in reduced effectiveness since the training data does not include difficult conditions ({\it e.g.} narrations, voice overs, overlapping sounds, challenging illumination, non-speech mouth motion). Recent efforts using heuristics to mine web video with imputed labels \cite{owens2018audio, ephrat2018looking, afouras18conversation} are similarly limited.

The approaches in the literature that directly tackle active speaker detection \cite{chakravarty2015who, li2005cross, cutler2000look, stefanov2016look} all present evaluations on task-specific datasets of less than an hour. In contrast, the AVA-ActiveSpeaker dataset provides several distinct benefits. This dataset contains realistic video with a wide diversity of recording conditions (background noise, illumination, {\it etc.}), speaker demographics, and temporally dense labels for each face. In addition, these are added to AVA's preexisting action and speech activity labels to further enrich the dataset, enabling cross-task analysis.

\textbf{Datasets:}
A number of efforts around developing audiovisual datasets with ground-truth labels are related. Early corpora were designed for speech (digit) recognition with high-resolution, frontal facing speaking faces, such as CUAVE~\cite{patterson2002cuave} and AVTIMIT~\cite{hazen2004segment}. The UT-CRSS-4EnglishAccent corpus for voice activity detection \cite{tao2017bimodal} and the AVDIAR corpus \cite{gebru2017audio} for diarization are recent corpora, but are limited in subject diversity and recording conditions. Meetings data \cite{anguera2006robust, mccowan2005ami} contain spontaneous speech, and  non-frontal and occluded faces. However, not all datasets contain video, or they have limited diversity due to high collection cost. Also, speaker labels are not associated with a visual person. Broadcast news corpora \cite{galliano2009ester, zelenak2012speaker} contain a larger speaker diversity fixed genre, due to being recorded in-studio. Datasets derived from movies and TV shows are also popular but are rarely more than a few hours and limited to a few shows and characters. For example, the REPERE corpus~\cite{giraudel2012repere} is 3 hours, Ren {\em et al}~\cite{ren2016look} evaluate on a single TV show with 5 characters, Everingham {\em et al}\cite{everingham2006buffy} use 2 episodes, Hu {\em et al}~\cite{hu2015deep} use 3 hours from 2 TV shows. 

In contrast, AVA-ActiveSpeaker contains $\sim40$K labeled face tracks totaling 38 hours, includes a variety of spoken languages,  is task-agnostic, and we had no influence over recording conditions, production or narrative structures.

\textbf{Multimodal approaches:}
Early multimodal approaches on speech (digit) recognition tasks projected the modalities into low dimensional subspaces that maximize the mutual information (MI) between the signals~\cite{slaney2001facesync,fisher2001learning,nock2003speaker,saenko2005visual}, with differences in feature representations and modelling paradigms. However, MI-based approaches do not perform well in unconstrained environments~\cite{vajaria2006audio}.

There exist a variety of approaches for conversation analysis in videos. While some use only visual information~\cite{stefanov2016look,stefanov2017vision}, many can be  considered ``multimodal": some incorporate audio and whole body information~\cite{vajaria2008exploring,friedland2009visual,hung2009speech}, some add localization signals obtained from a known microphone array configuration \cite{zhang2008boosting,gebru2018audio}, others utilize script and subtitle information~\cite{everingham2006buffy,everingham2009taking,bauml2013semi}. All of these perform late fusion with heuristics, typically assuming non-overlapping speech, and that the application is permissive to operating at high precision with low ($\sim30\%$) or unknown (since ground-truth is not available) recall.

Recent work jointly models audiovisual data, without handcrafted heuristics for late fusion, and two-tower neural network architectures (with a tower per modality). Hu {\em et al.}~\cite{hu2015deep} and Ren {\em et al.}~\cite{ren2016look} use audio and visual signals for automatic speaker naming. Hu {\em et al.} assumes non-overlapping speech, while Ren {\em et al.}~\cite{ren2016look}'s relaxes that assumption. Due to the absence of a dataset with dense, temporal labels, both evaluate over specified duration segments with model predictions treated as votes toward segment label prediction. In contrast, the densely labeled tracks in AVA-ActiveSpeaker allow more fine-grained analysis. Recent work in speech enhancement~\cite{chung2016out,gabbay2018visual,afouras18conversation,ephrat2018looking,owens2018audio} use two-tower neural networks; however, they also process the visual modality with pre-trained networks: using keypoints to isolate the mouth region \cite{chung2016out}, computing identity embeddings from detected faces ~\cite{ephrat2018looking}, and computing word-level lip reading embeddings~\cite{afouras18conversation}. These require additional technology and add computation. 

While the broader architecture choice for our audiovisual model resembles the general space of two-tower architectures, our models are trained on the detected face pixels directly and do not require any pre-trained networks. We present results on the AVA-ActiveSpeaker dataset with this multimodal network trained from scratch, and compare with visual-only and audiovisual models, as well as in various background noise conditions. Our experimental results describe the tradeoffs between model accuracy, latency and computational complexity. This modeling approach could be supplemented in the future with additional pre-trained embeddings if desired. In addition, it could be augmented by utilizing the rest of the AVA dataset movies to provide large amounts of unsupervised or semi-supervised data in conjunction with curriculum learning~\cite{bengio2009curriculum,graves2017curriculum} or reinforcement learning~\cite{levine2016endtoend,levine2016learning}.
   
\section{Dataset Construction}
\label{sec:labelinginterface}

Creating the AVA-ActiveSpeaker dataset consisted of four stages: video selection, label vocabulary definition, face track detection, and human annotation, producing the dense, spatio-temporal annotations that we refer to as the AVA-ActiveSpeaker dataset and which we will release publicly. A CSV file indicates face bounding boxes over time for each track and corresponding temporal labels. We provide examples of the spatio-temporal labeled tracks in Appendix \ref{sec:moredataset}, including examples of the actual CSV of track annotations. The dataset url, which will be used for download, is not included here due to the double-blind review process.


\vspace{-0.1in}
\paragraph{Video selection}
We labeled all available videos from v1.0 AVA dataset~\cite{gu2018ava}, each a continuous segment from minutes $15$ to $30$ from $188$ movies on YouTube. See Section 3 of \cite{gu2018ava} for details on the video selection process. While movies are not a perfect representation of {\it in-the-wild} data, this dataset was compelling for a few reasons: it contains movies from film industries around the world, leading to diversity in languages, recording conditions, and speaker demographics; the synchronized audio and visual streams enable development of joint audiovisual models; the dataset is already popular for action recognition and speech detection tasks, and enriching the dataset further provides the opportunity for cross-task modeling; finally, the structured narrative of movies provides the potential for extension to applications such as speaker diarization, a task made simpler by the presence of AVA-Speech labels, or plot or narrative structure analyses.

\vspace{-0.1in}
\paragraph{Label Vocabulary Definition}

The label vocabulary provided as part of the rating interface contains three options: {\em Not Speaking}, {\em Speaking and Audible}, and {\em Speaking but not Audible}.  The Speaking label is broken into two categories depending on the audio modality. {\em Speaking but not Audible} covers cases where someone may visually appear to be speaking, {\it e.g.} in the background, even though their speech is not audible in the soundtrack. This allows for fairer evaluation of visual-only approaches that should classify these instances as speech.

\vspace{-0.1in}
\paragraph{Face track generation}
As the example in Figure \ref{fig:rating_ui} shows, the speaker annotations are spatio-temporal and dense. People in the video are annotated by face bounding box tracks, and the synchronous audio waveform is shown above the video player. Since manual bounding box annotation is expensive, we use automatic face detection and tracking. Candidate faces are detected via a face detector similar to~\cite{li2015convolutional}, and tracked over time based on bounding box overlap and similarity, with gaps less than $0.2$ seconds within a track filled via Gaussian kernel smoothing of box corners. Tracks for labeling are required to be at least $1$ second long to provide sufficient context and remove spurious false positives, and no more than $10$ seconds to provide sufficient resolution on the audio waveform and prevent annotator fatigue. The occasional occurrence of merging of two identities into a single track or a false positive track generated by this process are both discarded by human annotators. This process produced $38,500$ tracks and $3.65$ million faces in the AVA-ActiveSpeaker dataset.

\vspace{-0.1in}
\paragraph{Active Speaker Annotation}
\label{sec:labeling-speechannotation}
The active speaker labels are generated by human annotators using the interface in Fig.~\ref{fig:rating_ui}. Each rating task contains a video clip with a bounding box around a single face, and the process is repeated for all visible faces in the clip. The activity timeline above the player depicts the audio waveform; it begins colorless and is filled with color coded labeled segments as the labeling progresses.  The timeline beneath the video depicts the face track time within the full video clip.  Both timelines may be clicked to seek to the corresponding time in the video. We provided detailed guidance to the annotators regarding various edge cases, which are discussed in Appendix \ref{sec:moredataset}.

\section{Labeled Dataset}
\label{sec:datasetstatistics}

\begin{figure*}[ht]
\begin{minipage}{0.32\linewidth}
  \centering
  \centerline{\includegraphics[width=\columnwidth]{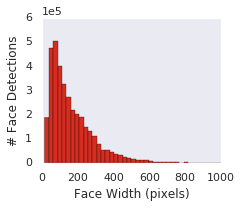}}
\end{minipage}
\hfill
\begin{minipage}{0.32\linewidth}
  \centering
  \centerline{\includegraphics[width=\columnwidth]{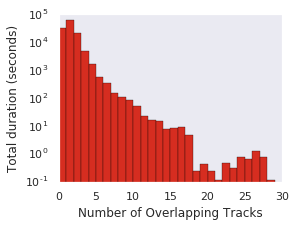}}
\end{minipage}
\hfill
\begin{minipage}{0.32\linewidth}
  \centering
  \centerline{\includegraphics[width=\columnwidth]{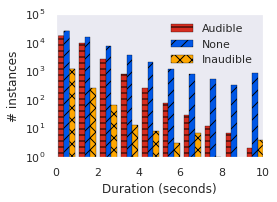}}
\end{minipage}
\caption{(a) Distribution of detected face widths that were labeled. (b) Distribution of total duration (log-scale) corresponding to the number of concurrently present faces. (c) Distribution of segment lengths for each label.}
\label{fig:pre_labeling_histogram}
\end{figure*}

Figure~\ref{fig:pre_labeling_histogram}a shows the distribution of face width over the detected face tracks. A significant portion of labeled faces are smaller than 100 pixels wide and likely to be challenging. Fig.~\ref{fig:pre_labeling_histogram}b shows the distribution by time of the number of concurrently present face tracks (note: y-axis is logarithmic). The higher end is crowded scenes with elevated levels of audio and visual activity, thereby making accurate predictions harder. Even with only two people, lively conversation and shot changes make classification difficult.

In total, the annotators labeled $\sim40K$ face tracks from 160 videos. Each track was manually labeled by three annotators, and the Fleiss' kappa \cite{fleiss1971measuring} value over the dataset was 0.72 indicating a high inter-annotator agreement. Most disagreements were near the temporal boundaries of speech segments, and due to perceptual differences. 

Table \ref{tab:data-stats} contains the summary statistics for the three labels. We see that the average duration of speaking face segments is surprisingly short. Although the average duration of continuous speech segments is higher (as reported in \cite{chaudhuri2018avaspeech}), speaking faces do not stay on screen through each utterance. In movies, the shot may pan the scene while speech is active, or visually cut to other scene elements; as a result, the face track is broken when the shot moves away from the speaker. We include video examples and discuss them in Appendix \ref{sec:moredataset}. As Figure \ref{fig:pre_labeling_histogram}c shows, the distribution of audible active speaker segment lengths appears to follow a power law.

\begin{table}[!t]
    \centering
    \begin{tabular}{|l|c|c|c|}
        \hline         
             Label & Time & \# Segments & Mean Duration \\\hline
         NS    & $28.10$ hours & $58,171$  & $1.74$ seconds \\\hline
         S\&A  & $9.46$ hours & $30,623$ & $1.11$ seconds \\\hline
         S\&NA  & $0.35$ hours & $1,547$  & $0.83$ seconds \\\hline
    \end{tabular}
    \caption{Aggregate statistics over the AVA-ActiveSpeaker dataset for the three labels: Not Speaking (NS), Speaking and Audible (S\&A), Speaking but Not Audible (S\&NA).}
    \label{tab:data-stats}
\end{table}

We analyzed the labeled tracks to determine how frequently active speakers overlapped in this dataset, {\em i.e.} multiple people were labeled as speaking at the same instant. While multiple co-occurring active speakers are  uncommon at $\sim500$ instances spanning $\sim3$ minutes, the number of faces in a single frame labeled as ``Speaking and Audible" is as high as 9, and labeled as ``Speaking but not Audible" faces is as high as 22. These appear to occur in group contexts, such as choruses. 

We note that while the face detection system used to generate candidate tracks for labeling is state-of-the-art and remarkably robust to a variety of conditions, it is still not perfect, and the challenging conditions such as illumination and partial occlusion do result in some missed detections. Appendix \ref{sec:moredataset} contains screenshots of some examples.

Previous annotations released on the videos in the AVA dataset have contributed action recognition (associated with a bounding box entity) \cite{gu2018ava} and audio speech activity labels (not associated with specific entities) \cite{chaudhuri2018avaspeech}. In the following two subsections, we discuss the relationships between the active speaker labels that we produce and those label sets.

\begin{figure*}[t]
\begin{minipage}{0.35\linewidth}
  \centering
  \centerline{\includegraphics[width=\columnwidth]{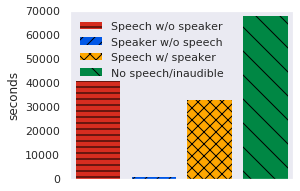}}
\end{minipage}
\hfill
\begin{minipage}{0.3\linewidth}
  \centering
  \centerline{\includegraphics[width=\columnwidth]{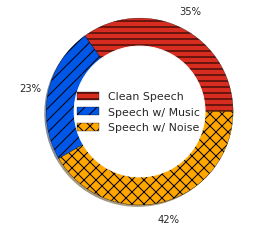}}
\end{minipage}
\hfill
\begin{minipage}{0.3\linewidth}
  \centering
  \centerline{\includegraphics[width=\columnwidth]{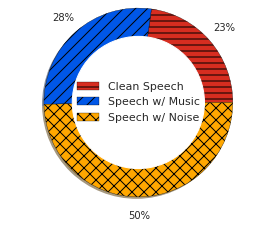}}
\end{minipage}
\caption{(a) Intersection between speech activity and active speaker labels. (b) Distribution of noise conditions when a speaker is visible. (c) Distribution of noise conditions when a speaker is not visible.}
\label{fig:speech-type-breakdown}
\vspace{-2mm}
\end{figure*}

\subsection{AVA action labels}
\label{subsec:ava_action_labels}

The AVA corpus \cite{gu2018ava} was originally released for visual action recognition research. A pair of labels--- ``talk-to" and ``sing-to"--- in the set of actions in AVA are relevant to active speaker detection focused exclusively in this work. There are two key differences in the processes to collect action labels and active speaker detection labels. First, the action labeling process was exclusively visual, whereas in our work the annotators had audio available. Second, the action labels were applied to the single middle frame of a 3-second context, thus providing annotators with limited context, whereas our labeling process makes entire tracks available to annotators resulting in densely annotated temporal label segments over the entire track. We computed the overlap between these AVA action labels and the active speaker labels in this dataset. Since AVA action labels are provided at a 1-second granularity and are not densely labeled, we can only determine how often the action label co-occurs with each active speaker labels (shown in Table \ref{tab:label-co-occur}, but cannot compute a similar number in the other direction.

\begin{table}[!t]
    \centering
    \begin{tabular}{|l|c|c|}
        \hline         
             Label & ``talk-to" & ``sing-to" \\\hline
         Not Speaking    & 17.05\% & 12.95\% \\\hline
         Speaking \& Audible  & 81.09\% & 71.07\% \\\hline
         Speaking but Not Audible  & 1.86\% & 15.98\% \\\hline
    \end{tabular}
    \vspace{0.1in}
    \caption{Co-occurrence of a pair of AVA action labels with the active speaker labels.}
    \label{tab:label-co-occur}
\end{table}

Table \ref{tab:label-co-occur} indicates that the annotators for the action labels and the active speaker disagree quite a bit. An examination of a sample of instances where the ``talk-to" labels appeared as ``Not Speaking" indicates many are near (but outside) the beginning or end of an actual speaking segment, while the rest are distributed over harder negatives, such as laughing, crying, yawning, etc. The correspondence with ``Speaking but not Audible" labels are as one would expect, covering instances where access to the audio would have allowed determining audibility.

The ``sing-to" labels corresponding to inaudible or not speaking labels occur in clearly musical contexts, but also show similar sources of confusion; {\it e.g.} words mouthed by a dancer, or a conductor, or frames near (but outside) boundaries of singing segments, where audio would have helped determine if someone was singing or not. We provide some illustrative examples in Appendix \ref{sec:moredataset}.

\subsection{AVA speech activity labels}
\label{subsec:avaspeechactivity}
Previous work has made speech activity labels - occurrence speech without any attribution to a specific visual entity - available on the AVA dataset \cite{chaudhuri2018avaspeech}. The labels released as part of this work extends this further to enable explicit attribution of speech to a visible face, when possible. Since the labeling process relies on faces, it will not cover cases where the speaker's face was not visible ({\it e.g.} offscreen speaker, back to the camera). So we do not expect to be able to attribute all the speech to visible speaking entities. Here, we quantify the proportion of speech activity that has been explicitly labeled with active speakers.

The speech activity labels indicate whether speech activity was present in the video at each instant in time. There are four speech activity types: ``No Speech", ``Clean Speech", ``Speech with Music", ``Speech with Noise". We combine the three different speech types into a single label and consider the binary condition: speech is occurring or not. Computing the overlap between Speech Activity and Active Speaker labels allows us to consider the following four cases:

\begin{enumerate}
    \item \textbf{Speech without Speaker}: Duration of speech heard, but not attributed to a visible speaker.
    \vspace{-2mm}
    \item \textbf{Speaker without Speech}: Duration of visible speaker, but no corresponding Speech Activity label recorded. 
    \vspace{-2mm}
    \item \textbf{Speech with Speaker}: Duration of overlapping speech and visible speaker audibly speaking.
    \vspace{-2mm}
    \item \textbf{No Speech/No audible speaker}: Duration with no active speech and no audible active speaker.
\end{enumerate}

The left panel in Figure \ref{fig:speech-type-breakdown} shows the total durations for with the four cases. As expected, the speaker without speech case only occurs due to frame-level disagreements at segment boundaries. The cases where active speech is attributed to an active speaker and cases where it is not are particularly interesting, and provides an insight into production effects in the movies genre: it is significantly due to the use of artistic narrative devices; {\it e.g.} the camera pans the scene as the viewer knows the speaker from a previous shot, speech overlaid on a ``dream" sequence in the video; some other cases consist of contextual speech that can be considered ``background"; {\it e.g. in a crowded market scene}. 

Since the previously released speech activity labels contain information about background noise (whether the speech is clean, or there is background music or other noises) when speech is present, we plotted the distribution of the speech condition information for the ``Speech without Speaker'' and the ``Speech with Speaker'' cases, shown in the middle and right panels of Figure \ref{fig:speech-type-breakdown}. We notice that the proportion of ``Clean Speech" increases by 12\% in the ``Speech with Speaker'' case: since the shot makes an effort to focus on the speaker, it is reasonable that a larger proportion would contain clean speech.

\section{Multimodal active speaker detection}
\label{sec:approach}
As discussed in Section \ref{sec:introduction}, robust active speaker detection requires the joint analysis of the audio and visual modalities. While this can be done using a late fusion of predictions from state-of-the-art single-modality models, we expect that a joint model that optimizes on the end-task will improve performance while being more efficient by reducing modeling redundancies across the single-modality models, and the need to train individual models as well as the combination model.

For the active speaker detection task, we desire to learn a mapping from a face track and audio signal to the probability of the face speaking at each time instant. That is, we want to learn $\mathbf{p} = f(\mathbf{I}, \mathbf{a}; w)$, where $\mathbf{I}=[\mathbf{I}_1, \mathbf{I}_2, \ldots, \mathbf{I}_N]$ is a track of face thumbnails, $\mathbf{a}=[a_1, a_2, \ldots, a_T]$ is a waveform (or a frequency-domain representation of the waveform), $\mathbf{p}=[p_1, p_2, \ldots, p_N]$ is the sequence of speech probabilities, and $w$ is the parameters to be trained to determine the mapping.  The function $f$ can be realized by any regressor, including a deep neural network.

\begin{figure}[t]
  \centering
  \includegraphics[width=\linewidth]{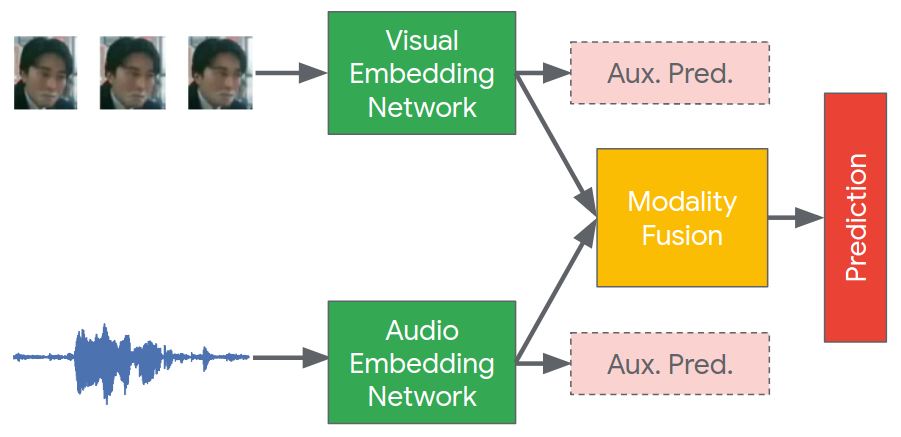}
  \caption{End-to-end multimodal active speaker detection framework.}
  \label{fig:asd_framework}
  \vspace{-2mm}
\end{figure}

In practice, we decompose $f$ (as shown in Fig.~\ref{fig:asd_framework}) into a set of DNNs to be jointly trained. That is, $f(\mathbf{I}, \mathbf{a}; w) = p(e_a(\mathbf{a}; w_a), e_v(\mathbf{I}; w_v); w_p)$, where $e_a \in \mathbb{R}^d$ is an audio embedding network, $e_v \in \mathbb{R}^d$ is a visual embedding network, and $p$ is a prediction network that fuses the low dimensional audio and visual embeddings. The audio or visual networks could be initialized with pre-trained networks, if desired, but in this work, we train from scratch with the pixels input directly to the visual network and the Mel-spectrogram representation of the audio is input to the audio network.

During training, each training example $i$ contains a sequence of face images $\mathbf{I}_i=[\mathbf{I}_{i1}, \mathbf{I}_{i2}, \ldots \mathbf{I}_{iN}$ sampled at $20$FPS, a corresponding audio representation $\mathbf{a}_i=[a_{i1}, a_{i2}, \ldots, a_{iT}]$ at $100$FPS, where $T = N / 20 * 100$, and a sequence of ground-truth labels $\mathbf{y}_i = [y_{i1}, \ldots, y_{iN}]$, where $y_{ij} = 1$ if the face $\mathbf{I_{ij}}$ is speaking and $y_{ij} = 0$ otherwise.  We  define the loss function $l(w)$ as a cross entropy loss between the predictions and labels:

\begin{equation}
l(w) = -\sum_j y_{ij} \log (p_{ij}) + \lambda \|w\|^2,
\label{eq:loss}
\end{equation}

where $\lambda$ is a regularization hyperparameter.  Furthermore, to encourage the prediction network to make use of both the audio and visual embeddings, we add independent auxiliary classification networks on each modality with corresponding cross entropy loss.  Our final loss is then a combination of all terms:

\begin{equation}
l(w) = L_{av} + \lambda_a L_a + \lambda_v L_v,
\end{equation}

where $L_{av}$ is Eq.~\ref{eq:loss}, $L_a$ and $L_v$ are the cross entropy of audio-only and visual-only networks, and $\lambda_a = \lambda_v = 0.4$ places a lower weight on the individual modality performance.

A wide variety of network choices can be explored for each network, but that is beyond the scope of this work. In particular, we expect that reasonable choices for the network will show that the joint audiovisual (AV, henceforth) models can significantly improve over visual-only (V, henceforth; and A for audio-only) models, and that these improvements hold up across a variety of conditions. While more sophisticated models will undoubtedly further push performance, we expect that the delta between the V and AV models will remain.

For the A and V networks, $e_a(\mathbf{a}; w_a)$ and $e_v(\mathbf{I}; w_v)$, we use a CNN employing the depthwise separable technique introduced by MobileNets~\cite{howard2017mobilenets}; our implementation uses fewer layers than the canonical MobileNet (since we use smaller images), and does not increase the number of $1 \times 1$ filters with network depth (to prevent overfitting on the relatively smaller dataset compared to ImageNet). Details of the network are in Table.~\ref{tab:cnn-structure}.

The input to the visual network is a stack of $M$ consecutive $128 \times 128$ grayscale face thumbnails.  By varying $M$, we can explore the effect of temporal information in the decision making process.  The Mel-spectrogram input to the audio network is $64 \times 48 \times 1$ and is computed over the preceding $0.5$seconds of audio, using a $25$ms analysis window.

\begin{table}[t]
\small
\centering
\begin{tabular}{l|l|l}
Type / Stride & Filter Shape & Input Size \\ \hline
Conv / s2    & $3 \times 3 \times M \times 32$     & $128 \times 128 \times M$ \\ \hline
Conv dw / s1 & $3 \times 3 \times 32$ dw           & $64 \times 64 \times 32$ \\
Conv / s1    & $1 \times 1 \times 32 \times 64$ dw & $64 \times 64 \times 32$ \\ \hline
Conv dw / s2 & $3 \times 3 \times 32$ dw           & $64 \times 64 \times 64$ \\
Conv / s1    & $1 \times 1 \times 32 \times 64$ dw & $32 \times 32 \times 64$ \\ \hline
Conv dw / s2 & $3 \times 3 \times 32$ dw           & $32 \times 32 \times 64$ \\
Conv / s1    & $1 \times 1 \times 32 \times 64$ dw & $16 \times 16 \times 64$ \\ \hline
Conv dw / s2 & $3 \times 3 \times 32$ dw           & $16 \times 16 \times 64$ \\
Conv / s1    & $1 \times 1 \times 32 \times 64$ dw & $8 \times 8 \times 64$ \\ \hline
Conv dw / s2 & $3 \times 3 \times 32$ dw           & $8 \times 8 \times 64$ \\
Conv / s1    & $1 \times 1 \times 32 \times 64$ dw & $4 \times 4 \times 64$ \\ \hline
Conv dw / s2 & $3 \times 3 \times 32$ dw           & $4 \times 4 \times 64$ \\
Conv / s1    & $1 \times 1 \times 32 \times 64$ dw & $2 \times 2 \times 64$ \\ \hline
Avg Pool / s1 & Pool $2 \times 2$                  & $2 \times 2 \times 64$ \\ \hline
FC / s1      & $64 \times 128$                     & $1 \times 1 \times 64$ \\ \hline
\end{tabular}
\caption{Visual embedding network architecture. The audio embedding network only differs in the input sizes.}
\label{tab:cnn-structure}
\vspace{-2mm}
\end{table}

For the prediction network, we explore two possibilities in this work. (1) A \emph{static} model, where each set of $M$ consecutive face frames are trained independently; and (2) a \emph{recurrent} model, where state from the previous timestep contributes to the determination of the current timestep's label.  The static model consists of two fully connected layers of $128$-dim and $2$-dim followed by a softmax to convert to probabilities.  The recurrent model consists of two $100$-dim Gated Recurrent Units (GRU) (as GRUs outperformed LSTMs, in our experiments) followed by a $2$-dim fully connected and softmax layers.
\section{Evaluation and Analysis}
\label{sec:evaluation}

\begin{figure*}[!t]
  \centering
  \includegraphics[width=\linewidth]{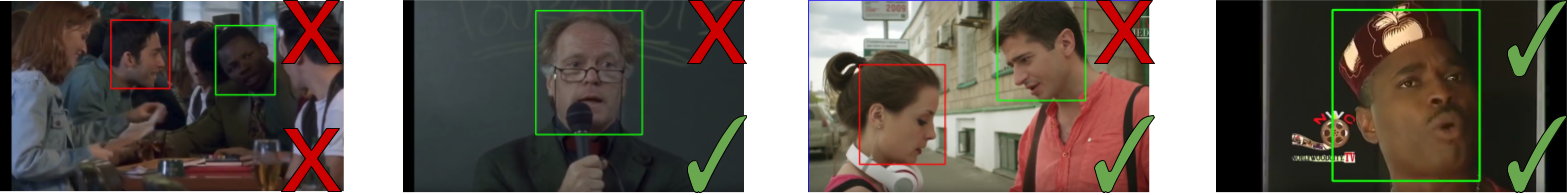}
  \caption{Sample ground truth speaking with frames: cross (predicted non-speech) and checkmark (predicted speech) for V Static f1 (top-right) and AV GRU f2 (bottom-right). The audiovisual model is able to handle occlusions and profile faces.}
  \label{fig:example-performance}
\end{figure*}

In this section, we present the results of evaluating the models described in Section \ref{sec:approach} on the test split of the AVA-ActiveSpeaker dataset, and present an extended analysis of model performance. All models were trained under the same conditions: we cropped the labeled tracks using a $3$-second ($60$-frame) sliding window with $1$-second overlap; we used an ADAGRAD optimizer with a learning rate of $2^{-6}$ for $10$ epochs.

\begin{figure}[!t]
  \centering
  \includegraphics[width=0.9\linewidth]{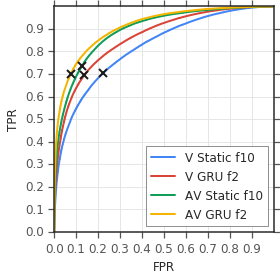}
  \caption{ROC curve for the best static and recurrent models with V and AV. `x' represents the balanced accuracy point.}
  \label{fig:roc}
\end{figure}

{\bf Metrics:} To compare model performances, we use area under the Receiver Operating Characteristic curve (auROC) as a holistic measure of performance. To slice a single model's performance across partitions of the data, we use balanced accuracy at a fixed $p=0.5$ threshold, chosen to remove bias from the number of positive labels in each partition. The fixed operating point ensures an equitable comparison.

{\bf Results:} Table~\ref{tab:results-summary} displays the overall results for the majority of experiments performed. We discuss them in detail below using the following abbreviations: visual-only as V, audio-visual as AV, and f$M$ as the number of stacked faces used in the visual embedding network.

\begin{table}[!t]
\centering
\begin{tabular}{l|rrr|rr}
 & \multicolumn{3}{c}{Static} & \multicolumn{2}{c}{GRU} \\
\# frames & V        & VV        & AV        & V        & AV \\ \hline
f1        & $0.68$   & $0.69$    & $0.83$    & $0.76$    & $0.86$ \\
f2        & $0.74$   & $0.73$    & $0.86$    & $0.86$    & $0.91$ \\
f3        & $0.77$   & $0.76$    & $0.87$    & $0.86$    & $0.92$ \\
f5        & $0.79$   & $0.79$    & $0.89$    & $0.86$    & $0.91$ \\
f7        & $0.81$   & $0.82$    & $0.89$    &           & \\
f10       & $0.82$   & $0.82$    & $0.90$    &           & \\
f15       & $0.82$   & $0.83$    & $0.90$    &           &  \\
\end{tabular}
\caption{auROC: Higher is better. GRU models start overfitting in $5$ frames or fewer, so we omit the f7, f10, f15.}
\label{tab:results-summary}
\end{table}

First, we look at visual-only model performances. We expect that an f1-V model should be able to detect speaking faces; \eg, humans can look at a photograph and guess if someone is speaking.  Indeed, ``V Static f1" performs better than random chance. Temporal information, however, should improve significantly over f$1$ models.  For all model types, performance jumps from f1 to f2.  Static models show continued improvement saturating around f10 or $0.5$sec of visual information. However, the GRU saturates at f2, which indicates that only short temporal motion is necessary to build visual embeddings, as the recurrent structure can use history from the beginning of the track.

AV models, however, should significantly outperform their corresponding V counterparts.  It is difficult for a V model to disambiguate speech from hard negatives that contain mouth motions, but an AV model can learn the relationship between the motions and audio signal for these challenging situations.  Indeed, static AV models show $>40\%$ reduction in error over V models, and the GRU models show $>30\%$ reduction.  Further, the AV static models require fewer visual frames before saturating, indicating that audio contributes significantly. 

To ensure the improvement from AV models is not simply due to twice the model parameters and embedding dimensions, we train a visual-visual (VV) model with two independent visual towers. VV is within hundredths of all V models, indicating that visual performance has indeed saturated, and that audio is key to performance improvement. Figure \ref{fig:roc} shows the full ROC curve for the best performing static and RNN settings for V and AV.

Table~\ref{tab:speech-breakdown} shows performance broken down by the background sound conditions, using information from the AVA-Speech labels~\cite{chaudhuri2018avaspeech}. Unlike audio-based speech detection performance reported in \cite{chaudhuri2018avaspeech}, V and AV models both show resilience to background sound. V models show similar performance regardless of the environment, while the AV model performance drops in the presence of background music (although still above V).

\begin{table}[!t]
\centering
\begin{tabular}{l|rrrr}
Model          & Clean    & Noise    & Music \\ \hline
V Static f10   & $76.3\%$ & $75.5\%$ & $75.4\%$ \\
AV Static f10  & $78.9\%$ & $78.2\%$ & $77.7\%$ \\
V GRU f2       & $79.2\%$ & $78.8\%$ & $78.1\%$ \\
AV GRU f2      & $81.9\%$ & $81.5\%$ & $79.8\%$ \\
\end{tabular}
\caption{Balanced accuracy across sound conditions.}
\label{tab:speech-breakdown}
\end{table}

Table~\ref{tab:face-size-breakdown} shows performance by face size, where small is $[0, 64)$ pixels, medium is $[64, 128)$, and large is $[128, \infty)$. As expected, both V and AV models perform better with larger faces; for V models, the false positive rate increases at a fixed threshold whereas it decreases for AV models.

\begin{table}[!t]
\centering
\begin{tabular}{l|rrrr}
Model          & Small    & Medium   & Large \\ \hline
V Static f10   & $70.5\%$ & $77.6\%$ & $82.2\%$ \\
AV Static f10  & $77.7\%$ & $85.6\%$ & $89.0\%$ \\
V GRU f2       & $74.5\%$ & $81.3\%$ & $84.2\%$ \\
AV GRU f2      & $78.7\%$ & $87.1\%$ & $89.4\%$ \\
\end{tabular}
\caption{Balanced accuracy by face size.}
\label{tab:face-size-breakdown}
\end{table}

We provide the full ROC plots for the different background sound and face size cases, as well as more examples and commentary on model performance in Appendix \ref{sec:moreresults}.

\subsection{ActivityNet Challenge}

\begin{table}[!t]
\centering
\begin{tabular}{l|rr}
           & V       & AV \\ \hline
Static f1  & $0.412$ & $0.656$ \\
Static f10 & $0.564$ & $0.738$ \\
GRU f2     & $0.711$ & $0.821$ \\
\end{tabular}
\caption{mAP of AVA ActiveSpeaker models on the held-out ActivityNet Challenge data.}
\label{tab:activity-net}
\end{table}

The AVA ActiveSpeaker dataset is part of the 4th ActivityNet challenge at CVPR 2019. Details on the task can be found in \href{http://activity-net.org/challenges/2019/tasks/guest_ava.html}{guest task B: Spatio-Temporal Action Localization}. The analysis of performance of models for this task is done on a held-out test set labeled separately for the challenge, which is available on the \href{https://research.google.com/ava/download.html#ava_active_speaker_download}{AVA ActiveSpeaker Download page}.

Table~\ref{tab:activity-net} reports the performance of a selected set of models computed through the ActivityNet evaluation server using the mean average precision metric. The models used here are the same ones whose performance was reported in Table \ref{tab:results-summary}.

\section{Conclusion}
\label{sec:conclusion}
This paper introduces the AVA-ActiveSpeaker dataset with dense, spatio-temporal annotations of spoken activity across $15$-min movie clips from the 160 videos in the AVA v1.0 dataset, creating the first publicly available, large-scale benchmark for the active speaker detection task. While the presence of face-associated speaking annotations already make this dataset interesting for various multimodal tasks, such as speaker identification, it also provides the opportunity for future work to develop the annotations further to extend to tasks such as speaker diarization, or even more holistic analyses tasks around plot and narrative structures. We also present a joint audiovisual modeling approach for the active speaker detection task, which reduces the errors in visual-only approaches by 36\%, and present an analysis of model performance across several conditions.

{\small
\bibliographystyle{ieee}
\bibliography{asd}
}

\newpage
\appendix
\part*{Appendix}










\appendix

Section \ref{sec:moredataset} contains additional information on the dense, spatio-temporal labels in the AVA-ActiveSpeaker dataset that we will release and the labeling process, and Section \ref{sec:moreresults} adds supplementary information related to model performance on this dataset.

\section{Dataset Information}
\label{sec:moredataset}

\begin{figure*}[!h]
  \centering
  \includegraphics[ height=3in]{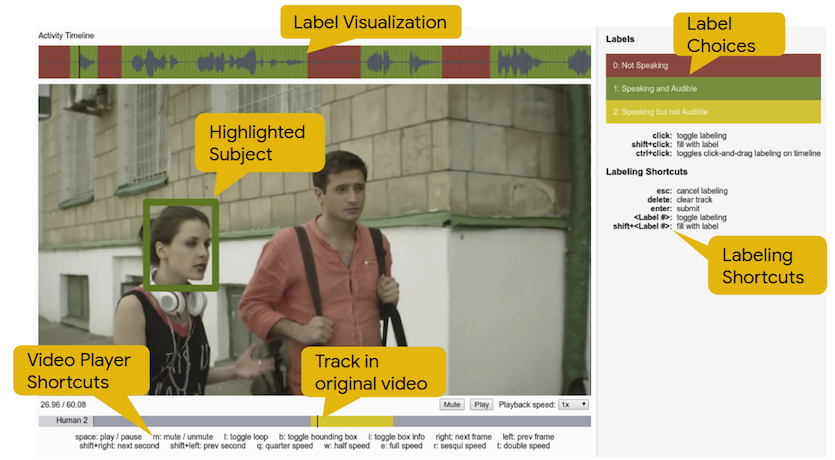}
  \caption{The annotation interface for AVA-ActiveSpeaker, with the interface components marked with yellow boxes. As rating progresses, the labels are overlaid on the audio waveform timeline - the snapshot here shows a fully labeled timeline. Raters can skip to any point in the video and the box around the subject's face corresponds to the color of the label at that instant. Raters can modify labels as necessary.}
  \label{fig:annotated-labeling}
  \vspace{-2mm}
\end{figure*}

\paragraph{Rating Interface:} A larger image of the rating interface (Figure \ref{fig:rating_ui} in the main paper) is in Figure \ref{fig:annotated-labeling}, with UI components highlighted. Raters can jump to any point in the track using the timeline, modify the label, view the track in the context of the video, use keyboard shortcuts to view at different speeds.

\paragraph{Videos with labels visualized:}
A pair of clips with ActiveSpeaker labels visualized are in the video files linked below. The bounding box color indicates the label at each instant: red for ``Not Speaking", green for ``Speaking and Audible", and yellow for ``Speaking but Not Audible".

\begin{enumerate}
\itemsep0em 
    \item \href{https://youtu.be/AhHwyAKuVIE}{speaker-labeled-1}: A "conversation" between 3 people. Near the end of this clip, the spoken content is replaced with music as an illustrative example for the ``Speaking but not Audible" label.
    \item \href{https://youtu.be/MPdCmKWZcgw}{speaker-labeled-2}: Audible speech between the two participants. Note that one of them speaks (``Mark me") before entering the visual scene, and that portion of speech is not part of the labeled data, since the speaker was not visible.
\end{enumerate}

\begin{figure*}[h]
  \centering
  \includegraphics[width=\linewidth]{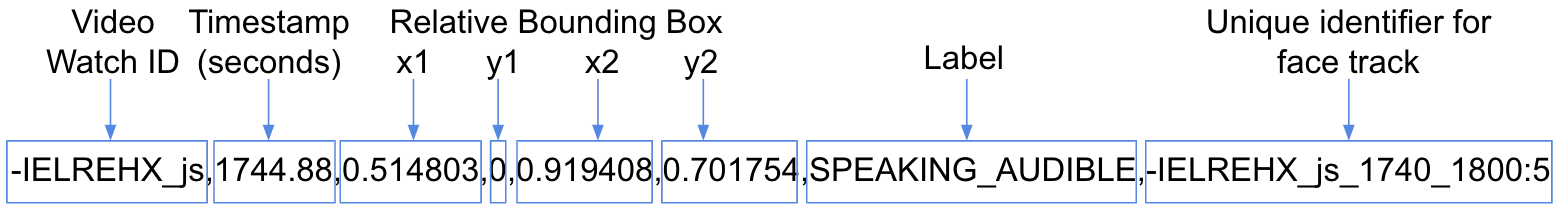}
  \caption{A single line illustrating the format of the released CSV data, with the annotations indicating the function of each of the eight comma separated values. ($x_1, y_1$) and ($x_2, y_2$) are the normalized locations of the top left and bottom right of the bounding box.}
  \label{fig:csv_format}
  \vspace{-2mm}
\end{figure*}

\paragraph{Release Data Format:}
Figure \ref{fig:csv_format} shows an example entry and the interpretation of the comma-separated values. The full CSV file will be available on the \href{https://research.google.com/ava/}{AVA website} soon.

\paragraph{Rating Guidance:}
Detailed instructions for a variety of cases were provided to raters. Table.~\ref{tab:speech_not_speech} summarizes guidance used to decide what should be considered speaking. Additionally, speaking faces are audible if raters could ascertain that the subject's speech was audible in the audio; clearly dubbed speech, speakers not heard due to overlapping sounds, or if the audio had music (or other sounds) overlaid for cinematic effect  should be labeled ``Speaking but not Audible''.  Raters had access to audio and video while making all decisions.

\newcommand\tabitem{\makebox[1em][r]{\textbullet~}}
\begin{table}[!h]
    \centering
    \begin{tabular}{|p{3.7cm}|p{3.7cm}|}
        \hline         
           \centering{Speaking} & Not Speaking \\\hline
            \tabitem Short utterances (\eg, ``Yes.'', ``Go.'' or ``Hmm'')&\tabitem Sighs, coughs, laughs, groans, grunts\\
            \tabitem Singing (with or without music) &\tabitem Mouthing along with music\\
            \tabitem Vocalized communication intent (\eg, scream to attract attention) &\tabitem Non-spoken communication (e.g. gesturing, waving)\\
            \tabitem Fillers (``um'', ``ah'')&\tabitem Humming \\
        \hline
    \end{tabular}
    \caption{Rater guidance indicating to identify speaking instances.}
    \label{tab:speech_not_speech}
\end{table}

\paragraph{AVA Action Label Differences}
The choice of the AVA corpus allows us to correlate ActiveSpeaker labels with previously-released  action labels \cite{gu2018ava} (where annotators only had access to the visual modality) and speech activity labels \cite{chaudhuri2018avaspeech}. Table 2 notes that $\sim17\%$ of ``talk-to" labels and $\sim13\%$ of ``sing-to" in the AVA actions dataset were errors that do not correspond to someone speaking (audible or inaudible) in the ActiveSpeaker dataset. Below, we describe three conditions in the AVA actions dataset discovered by cross-checking the action labels with our speaker labels.

\begin{figure*}[!t]
\begin{minipage}{0.31\linewidth}
  \centering
  \centerline{\includegraphics[width=\columnwidth,height=1.28in]{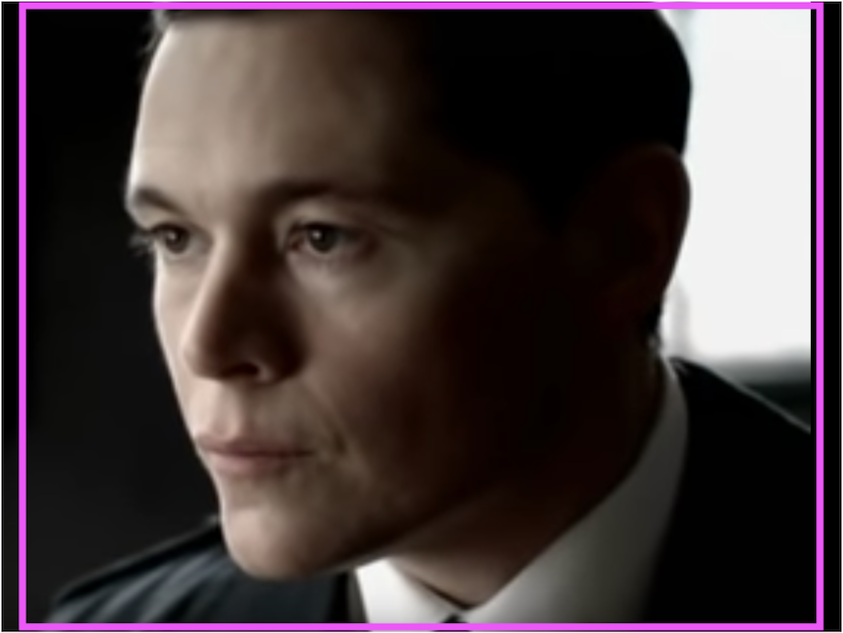}}
\end{minipage}
\hfill
\begin{minipage}{0.31\linewidth}
  \centering
  \centerline{\includegraphics[width=\columnwidth,height=1.28in]{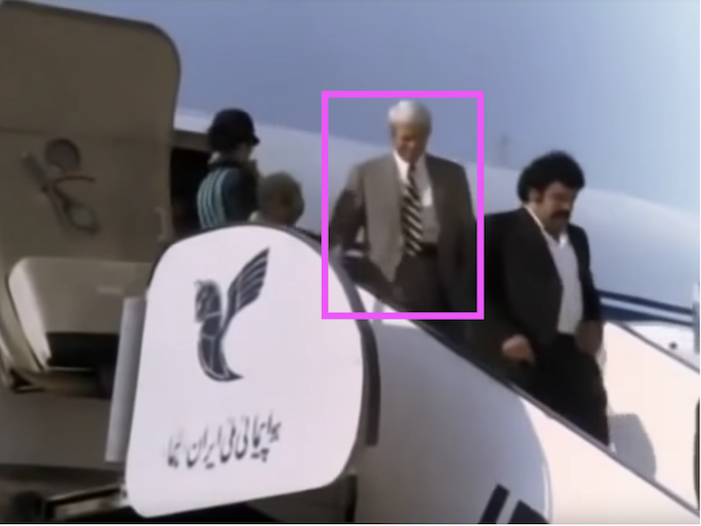}}
\end{minipage}
\hfill
\begin{minipage}{0.31\linewidth}
  \centering
  \centerline{\includegraphics[width=\columnwidth,height=1.28in]{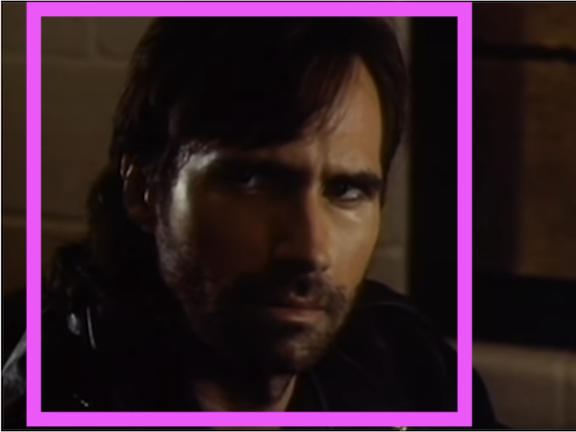}}
\end{minipage}
\caption{Frames with incorrect AVA action label ``talk-to" when the subject (in the pink bounding box) wasn't speaking.}
\label{fig:talk-to-nospeech}
\end{figure*}

\begin{figure*}[!t]
\begin{minipage}{0.31\linewidth}
  \centering
  \centerline{\includegraphics[width=\columnwidth,height=1.28in]{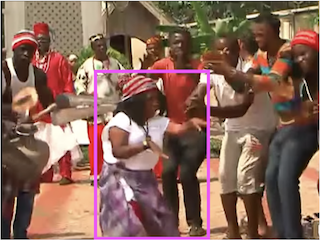}}
\end{minipage}
\hfill
\begin{minipage}{0.31\linewidth}
  \centering
  \centerline{\includegraphics[width=\columnwidth,height=1.28in]{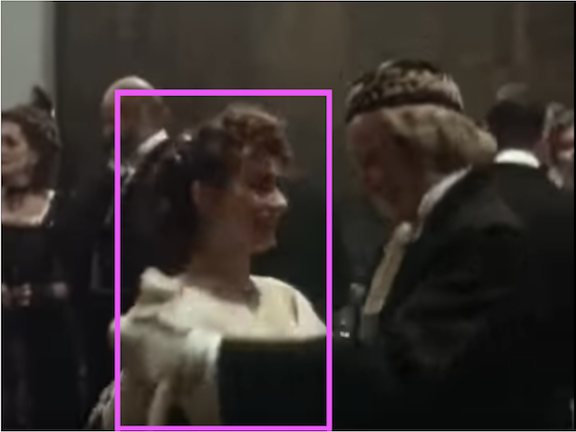}}
\end{minipage}
\hfill
\begin{minipage}{0.31\linewidth}
  \centering
  \centerline{\includegraphics[width=\columnwidth,height=1.28in]{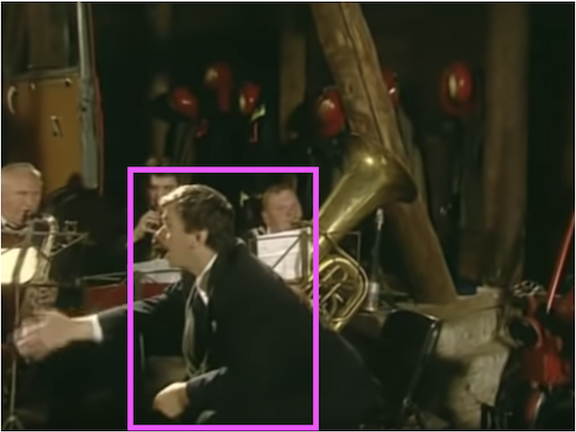}}
\end{minipage}
\caption{Frames of ``sing-to" AVA action labels where the subject was not actually singing. All examples have a musical context. (Left \& Center) The subjects were dancing to music. (R) Subject is a conductor directing the musicians, with emphatic mouth (and body) motions.}
\label{fig:sing-to-errors}
\end{figure*}

\begin{itemize}
\itemsep0em 
    \item {\bf Incorrect ``talk-to" labels}: Figure \ref{fig:talk-to-nospeech} shows instances where a face was not speaking but was labeled ``talk-to". These usually occur just outside the boundaries of a speaking segment. The AVA action annotators were misled by the lack of audio while labeling and proximity to speaking frames.
    \item {\bf Inaudible ``talk-to" labels}: Here, true semantics cannot be ascertained from visual-only labeling: the face looks like its speaking, but it is not heard in the audio which often contains overlaid music for cinematic effect. \eg the section starting at 23 seconds in \href{https://youtu.be/AhHwyAKuVIE?t=24}{speaker-labeled-1}. 
    \item {\bf Incorrect ``sing-to" labels}: Figure \ref{fig:sing-to-errors} shows cases where the specified entity was not actually singing; once again, the absence of audio makes it hard for the annotators to accurately determine when someone is singing or just lip-syncing along with the music.  
\end{itemize}

\begin{figure*}[!t]
\begin{minipage}{0.31\linewidth}
  \centering
  \centerline{\includegraphics[width=\columnwidth,height=1.28in]{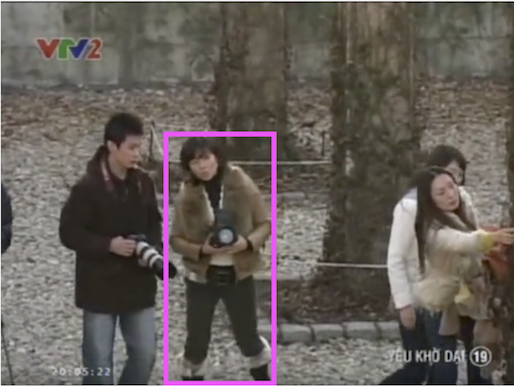}}
\end{minipage}
\hfill
\begin{minipage}{0.31\linewidth}
  \centering
  \centerline{\includegraphics[width=\columnwidth,height=1.28in]{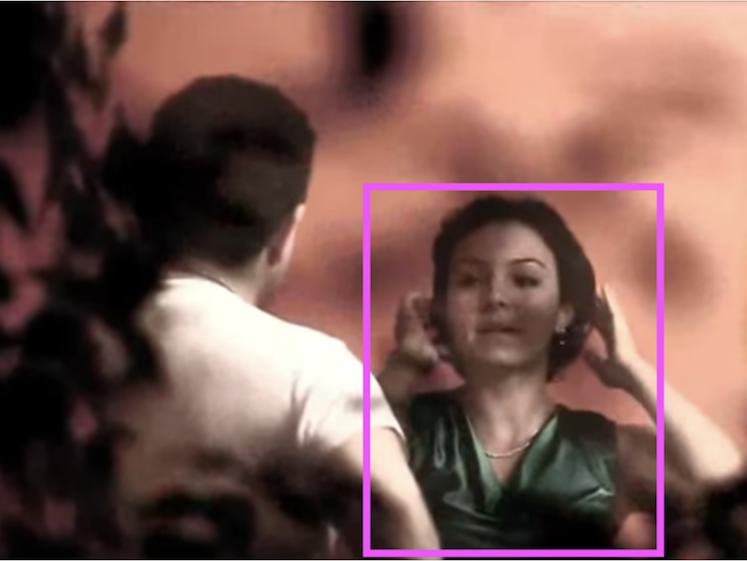}}
\end{minipage}
\hfill
\begin{minipage}{0.31\linewidth}
  \centering
  \centerline{\includegraphics[width=\columnwidth,height=1.28in]{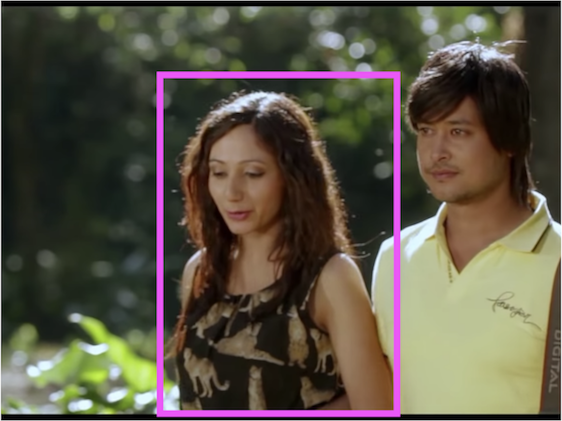}}
\end{minipage}
\caption{Frames with ``talk-to" AVA action label where the subject appears to be speaking but their speech is not in the audio track, instead overlaid with music or sound effects for cinematic effect.}
\label{fig:talk-to-inaudible}
\end{figure*}

\begin{figure*}[!t]
\begin{minipage}{0.31\linewidth}
  \centering
  \centerline{\includegraphics[width=\columnwidth,height=1.25in]{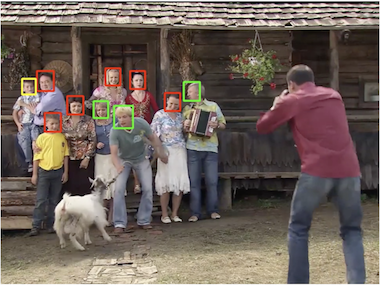}}
  \centerline{}\medskip
\end{minipage}
\hfill
\begin{minipage}{0.31\linewidth}
  \centering
  \centerline{\includegraphics[width=\columnwidth,height=1.25in]{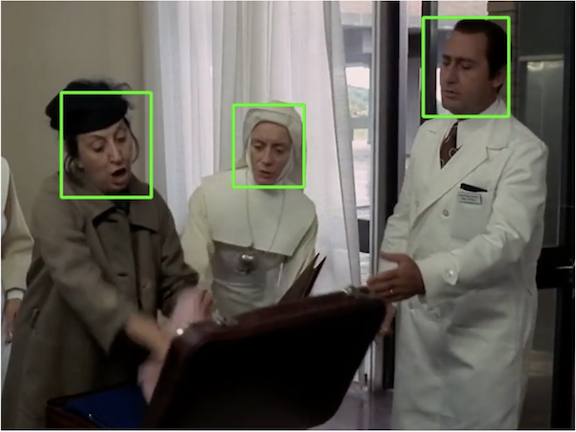}}
  \centerline{}\medskip
\end{minipage}
\hfill
\begin{minipage}{0.31\linewidth}
  \centering
  \centerline{\includegraphics[width=\columnwidth,height=1.25in]{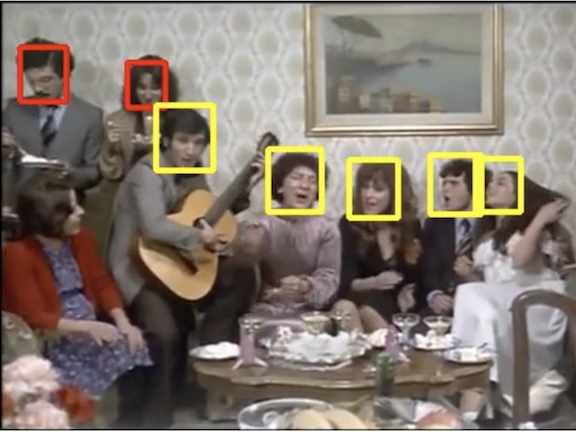}}
  \centerline{}\medskip
\end{minipage}
\caption{Visualizations of overlapping speaker instances. A green bounding box represents speaking and audible, yellow represents speaking and inaudible, red represents not speaking.}
\label{fig:overlapping_speakers}
\end{figure*}

\begin{figure*}[!t]
\begin{minipage}{0.31\linewidth}
  \centering
  \centerline{\includegraphics[width=\columnwidth,height=1.25in]{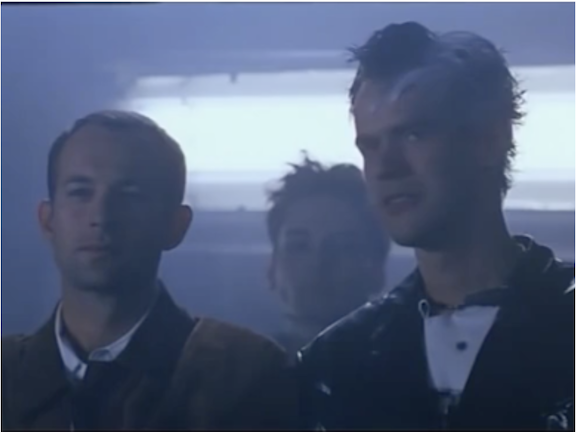}}
  \vspace{0.1cm}
  \centerline{}\medskip
\end{minipage}
\hfill
\begin{minipage}{0.31\linewidth}
  \centering
  \centerline{\includegraphics[height=1.25in, width=\columnwidth]{images/eg-many-faces}}
  \vspace{0.1cm}
  \centerline{}\medskip
\end{minipage}
\hfill
\begin{minipage}{0.31\linewidth}
  \centering
  \centerline{\includegraphics[width=\columnwidth,height=1.25in]{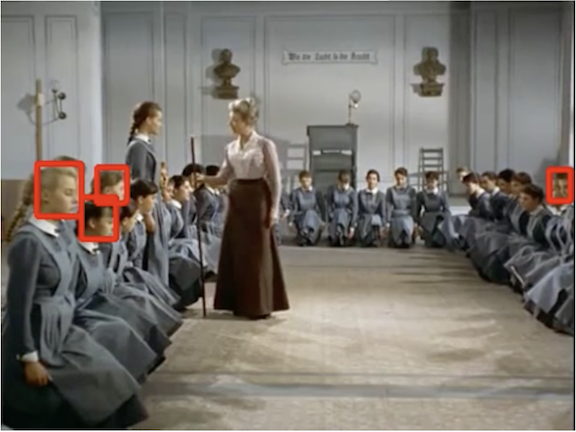}}
  \vspace{0.1cm}
  \centerline{}\medskip
\end{minipage}
\caption{(Left) Missed face detections in backlit conditions; (Center) A few missed detections in the back of a crowd; (Right) Missed detections as faces get smaller.}
\label{fig:bad-detections}
\end{figure*}

\paragraph{AVA Speech Label Differences:} Section \ref{subsec:avaspeechactivity} of the main paper discusses the AVA-ActiveSpeaker labels in the context of the previously released speech activity labels \cite{chaudhuri2018avaspeech}. One surprising observation was that a significant amount of time labeled as containing speech in AVA-Speech did not have an  active speaker at that instant in AVA-ActiveSpeaker (Figure \ref{fig:speech-type-breakdown} in main paper). A sampling of such cases shows that the shot often does not directly focus on the active speaker in movies for cinematic effect; viewers have enough context and voice recognition to know the speaker, anyway. Background music with vocals are labeled as containing speech, but naturally cannot be associated with a visible speaker. \href{https://youtu.be/oD_wxyTHJ2I?start=1201&end=1241}{This clip} from the AVA dataset contains an illustrative example. It begins with off-screen speech, transitions to a scene with background music with vocals, and finally the riders of the car speak with their faces not visible.

\vspace{-0.1in}
\paragraph{Overlapping Speakers:}
The detailed per-person labels allow us to identify segments with overlapping speakers, potentially interesting for audiovisual speech separation efforts. Figure \ref{fig:overlapping_speakers} shows snapshots from such segments identified with the dense labels in AVA-ActiveSpeaker.

\vspace{-0.1in}
\paragraph{Missed Face Tracks:}
Section \ref{sec:labelinginterface} of the paper describes the automated process for obtaining the face tracks which form the basis of the labels in AVA-ActiveSpeaker. While the detection and tracking pipeline is state-of-the-art, the videos in this dataset contain a number of challenging cases such as crowded scenes, small faces, challenging lighting condition, partially occluded faces, {\em etc.}, in conjunction with the varied resolutions of the video itself, many of them lower than the production quality of movies and TV shows of today. This leads to some missed face track detection which are not labeled. Figure \ref{fig:bad-detections} contains a sample of instances where  faces were missed.

\begin{figure*}[!t]
	\centering
	\begin{tabular*}{\textwidth}{@{}c@{}c@{}c@{}c@{}}
		\includegraphics[width=.25\linewidth]{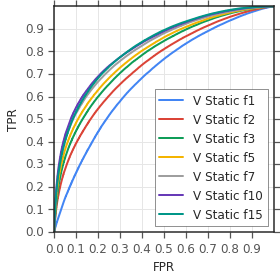} &
		\includegraphics[width=.25\linewidth]{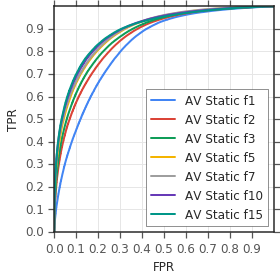} &
		\includegraphics[width=.25\linewidth]{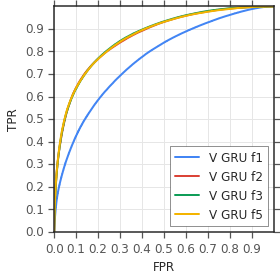} &
		\includegraphics[width=.25\linewidth]{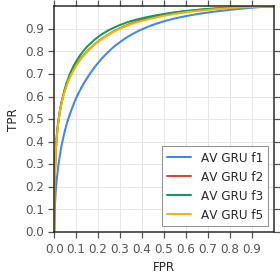} \\
	\end{tabular*}
	\caption{ROC curves for V and AV static and recurrent models, corresponding to Table \ref{tab:results-summary} from the paper.; From left to right, (a) static V models, (b) static AV models, (c) recurrent V models, (d) recurrent AV models (f2 , f3 and f5 are on top of each other for recurrent models). }
	\label{fig:all-roc-sweep}
\end{figure*}

\begin{figure*}[tbp!]
	\centering
	\begin{tabular*}{\textwidth}{@{}c@{}c@{}c@{}c@{}}
   	    \includegraphics[width=.25\linewidth, height=1.5in]{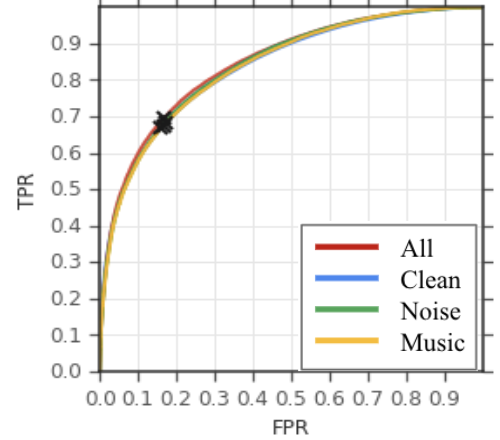} &
		\includegraphics[width=.25\linewidth, height=1.5in]{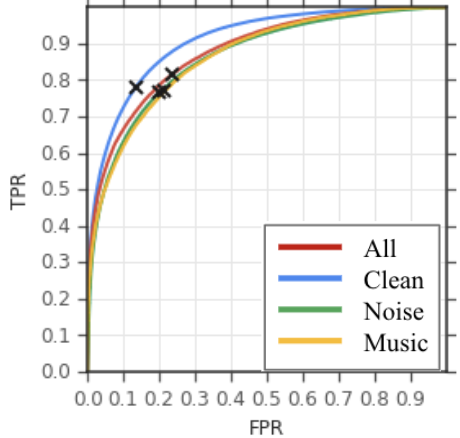} &
		\includegraphics[width=.25\linewidth, height=1.5in]{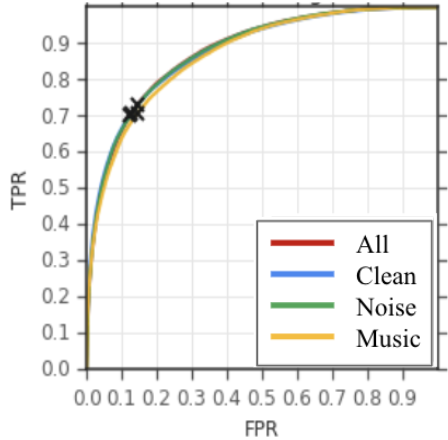} &
		\includegraphics[width=.25\linewidth,height=1.5in]{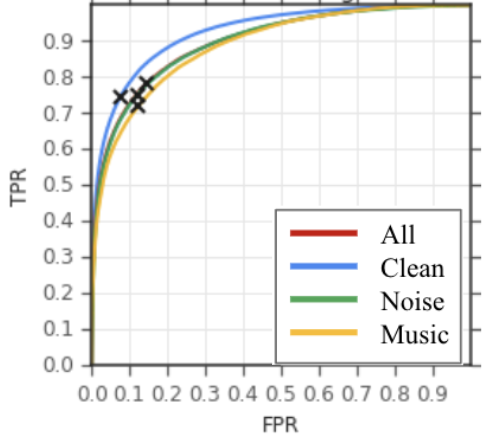} \\
	\end{tabular*}
	\caption{ROC curves partitioned by background sound determined by AVA-Speech labels. From left to right, (a) static V models, (b) static AV models, (c) recurrent V models, (d) recurrent AV models. The $\times$ represents the $p=0.5$ balanced accuracy point.}
	\label{fig:speech-sound-breakdown}
\end{figure*}

\begin{figure*}[!t]
	\centering
	\begin{tabular*}{\textwidth}{@{}c@{}c@{}c@{}c@{}}
		\includegraphics[width=.25\linewidth]{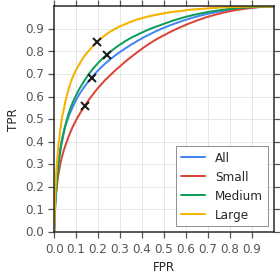} &
		\includegraphics[width=.25\linewidth]{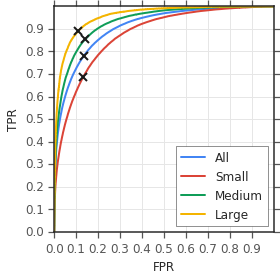} &
		\includegraphics[width=.25\linewidth,height=1.73in]{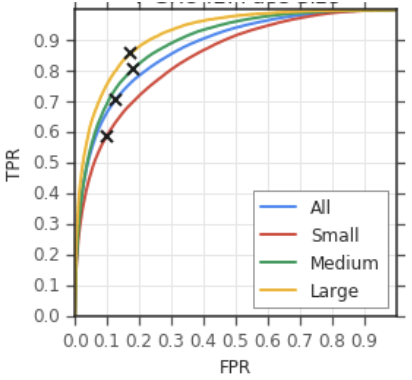} &
		\includegraphics[width=.25\linewidth,height=1.73in]{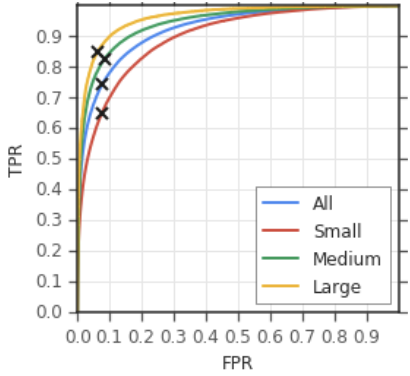} \\
	\end{tabular*}
	\caption{ROC curves for partitioned by face sizes, corresponding to Table 6 from the paper. From left to right, (a) static V models, (b) static AV models, (c) recurrent V models, (d) recurrent AV models. The $\times$ on each curve represents the $p=0.5$ balanced accuracy point.}
	\label{fig:speech-face-breakdown-roc}
\end{figure*}

\section{Supplementary Results}
\label{sec:moreresults}

We use the same abbreviations to denote model types here as in the main paper; V: visual-only model, AV: audiovisual model, GRU: gated recurrent unit models,  f$M$: number of frames in the stack input to the visual network. 

\vspace{-0.1in}
\paragraph{Model Comparison:}
Figure \ref{fig:all-roc-sweep} contains the full ROC curves for Table \ref{tab:results-summary} from the paper. In static models, performance keeps improving till $M=10$, and for each $M$, the corresponding AV curve is considerably better than V; AV-GRU is $\sim10\%$ better TPR than V-GRU and $\sim5\%$ better TPR than AV-static at $10\%$ FPR. The same pattern holds with recurrent models, although performance improvements saturate at 2 frames, indicating that only a short amount of history is needed.

\vspace{-0.1in}
\paragraph{Effect of background noise:}
 Figure~\ref{fig:speech-sound-breakdown} shows the full ROC curves for Table \ref{tab:speech-breakdown} from the paper. Unlike audio-based speech detectors' performance reported in \cite{chaudhuri2018avaspeech}, both V and AV models show resilience to background sound. While V models are not affected at all, AV models' performance slightly dips in overlapping music and noise, although still outperforming V models.

 \begin{figure*}[!t]
     \centering
 	\includegraphics[width=\linewidth]{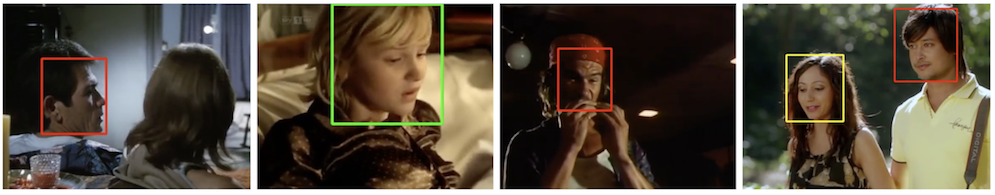}
 	\caption{Frames where V model predictions were incorrect, while AV-GRU were correct. AV models can use audio to know if speech is occurring to correct false positives from V when there is no speech.}
 	\label{fig:v-wrong-av-right}
 \end{figure*}
 
 \begin{figure*}[!t]
     \centering
 	\includegraphics[width=\linewidth]{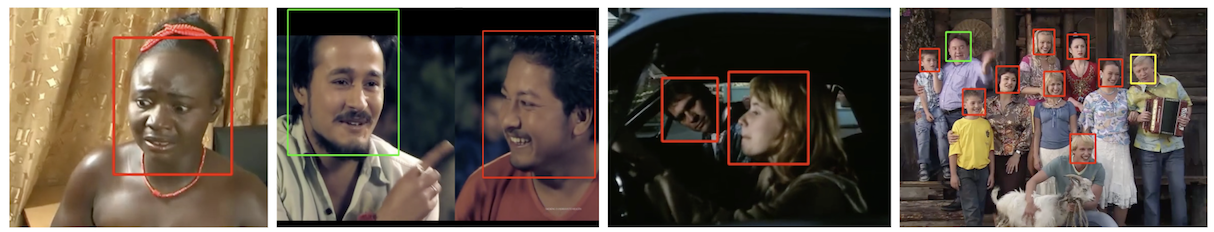}
 	\caption{Frames where AV-static-f10 models made errors, but AV-GRU-f2 got them right; GRU models seem to be capturing synchronization information beyond detecting speech.}
 	\label{fig:av-static-wrong-gru-right}
 \end{figure*}
 
  \begin{figure*}[!t]
     \centering
 	\includegraphics[width=\linewidth]{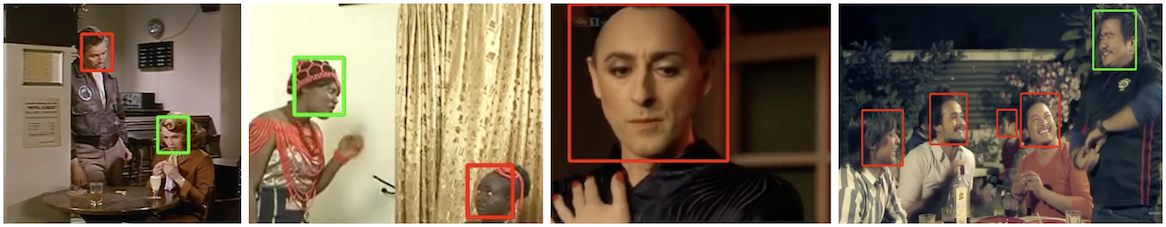}
 	\caption{Frames where AV-GRU-f2 made errors; Left 2: small faces; Right 2: challenging overlaid speaking while other faces made lip movements with vocalized sounds.}
 	\label{fig:av-gru-wrong}
 \end{figure*}

\vspace{-0.1in}
\paragraph{Effect of face size:} Figure ~\ref{fig:speech-face-breakdown-roc} shows ROC curves for Table \ref{tab:face-size-breakdown} of the paper, partitioned by face size: small ($< 64$ pixels wide), medium ($ > 64$, $< 128$ px) and large ($> 128$ px, larger than model input). AV models clearly outperform V models: for GRU, absolute improvement in TPR at $10\%$ FPR is $\sim10\%$ for small faces, $\sim15\%$ for medium, $\sim13\%$ for large.  The biggest difference in ``medium" suggests that this might be the sweet spot for the combined advantage of recurrence and AV: for smaller faces, the visual information is harder to leverage for all models, while for larger faces, visual information is enough for V to close the gap.

For AV models, FPR at the balanced accuracy point is nearly constant, while for V, they are more variable. For applications that mine data corresponding to speaking faces (for tasks like  synchronization \cite{chung2016out}, visual speech recognition \cite{shillingford2018lsvsr}, enhancement \cite{afouras18conversation}), these models can be deployed without needing additional calibration and hand-tuning.

\vspace{-0.1in}
\paragraph{Examples of model predictions:} Figure~\ref{fig:v-wrong-av-right} shows frames where V model made errors while AV-GRU-f2 were correct. AV models appear more robust to pan angles and profiles (left two panels), can use audio context to know speech isn't occurring (second from right), and are more robust to partial occlusions and motion around the face (right panel).

Figure~\ref{fig:av-static-wrong-gru-right} shows frames where AV static models were wrong but AV GRU got them right. Improvements from static to GRU within AV models appear to be driven by an enhanced ability to understand synchronization between the audio and visuals, even though the models were not explicitly trained for it. This makes the better models robust to noise in the audio domain (background music) as well as visual domain (partial occlusions).

Figure ~\ref{fig:av-gru-wrong} shows frames where AV-GRU models made the wrong prediction. Based on our sampling, there appear to be 2 clear modes of failure. One occurs when the faces are small and there is motion in more than one face (left 2 panels), where the model will pick both as speaking possibly due to not having a clear enough visual signal for who to associate the speech with. The other is when multiple sources are vocal but only one is speaking and others are,. \eg, laughing (right 2 panels). One way to alleviate some of these issues would be to explicitly add in augmentation at training time geared toward enabling the model to learn explicit synchronization.

\end{document}